\documentclass[runningheads]{llncs}
\usepackage{graphicx}

\usepackage{hyperref}
\usepackage{booktabs}
\usepackage{multirow}
\usepackage{array}
\newcommand{\tabincell}[2]{\begin{tabular}{@{}#1@{}}#2\end{tabular}}
\usepackage{amsmath,amssymb} 
\usepackage[table]{xcolor}
\mathchardef\mhyphen="2D
\usepackage{bm}
\usepackage{tikz}
\usepackage{comment}
\usepackage{amsmath,amssymb} 
\usepackage{color}
\usepackage[ruled,vlined,linesnumbered]{algorithm2e}
\usepackage{enumerate}
\usepackage[misc]{ifsym}

\begin{document}
\pagestyle{headings}
\mainmatter
\def\ECCVSubNumber{343}  

\title{DID-M3D: Decoupling Instance Depth for Monocular 3D Object Detection} 

\titlerunning{DID-M3D: Decoupling Instance Depth for Monocular 3D Object Detection}
\authorrunning{L. Peng et al.}
%


\author{Liang Peng\inst{1,2}, Xiaopei Wu\inst{1}, Zheng Yang\inst{2}, Haifeng Liu\inst{1}, and  Deng Cai\inst{1,2} \textsuperscript{\Letter}
}
\institute{
State Key Lab of CAD\&CG, Zhejiang University, China  \\
\email{\{pengliang, wuxiaopei, haifengliu\}@zju.edu.cn \, dengcai@cad.zju.edu.cn} \\
\and Fabu Inc., Hangzhou, China\\
\email{yangzheng@fabu.ai} 
}

\maketitle

\begin{abstract}
	Monocular 3D detection has drawn much attention from the community due to its low cost and setup simplicity.
	It takes an RGB image as input and predicts 3D boxes in the 3D space.
	The most challenging sub-task lies in the instance depth estimation.
	Previous works usually use a direct estimation method. 
	However, in this paper we point out that the instance depth on the RGB image is non-intuitive.
	It is coupled by visual depth clues and instance attribute clues, making it hard to be directly learned in the network.
	Therefore, we propose to reformulate the instance depth to the combination of the instance visual surface depth \textbf{(visual depth)} and the instance attribute depth \textbf{(attribute depth)}.
	The visual depth is related to objects' appearances and positions on the image.
	By contrast, the attribute depth relies on objects' inherent attributes, which are invariant to the object affine transformation on the image.
	Correspondingly, we decouple the 3D location uncertainty into visual depth uncertainty and attribute depth uncertainty.
	By combining different types of depths and associated uncertainties, we can obtain the final instance depth.
	Furthermore,  data augmentation in monocular 3D detection is usually limited due to the physical nature, hindering the boost of performance. 
	Based on the proposed instance depth disentanglement strategy, we can alleviate this problem.
	Evaluated on KITTI, our method achieves new state-of-the-art results, and extensive ablation studies validate the effectiveness of each component in our method.
	The codes are released at \href{https://github.com/SPengLiang/DID-M3D}{https://github.com/SPengLiang/DID-M3D}.
	
\end{abstract}

\section{Introduction}

	Monocular 3D object detection is an important topic in the self-driving and computer vision community.
	It is popular due to its low price and configuration simplicity.
	Rapid improvements \cite{Mono3D,MonoRUn,Monodle,peng2021lidar,DFR,GUPNet} have been conducted in recent years.
	A well-known challenge in this task lies in instance depth estimation, which is the bottleneck towards boosting the performance since the depth information is lost after the camera projection process.
	
\begin{figure}[t]
\centering 
				\includegraphics[width=1.0\linewidth]{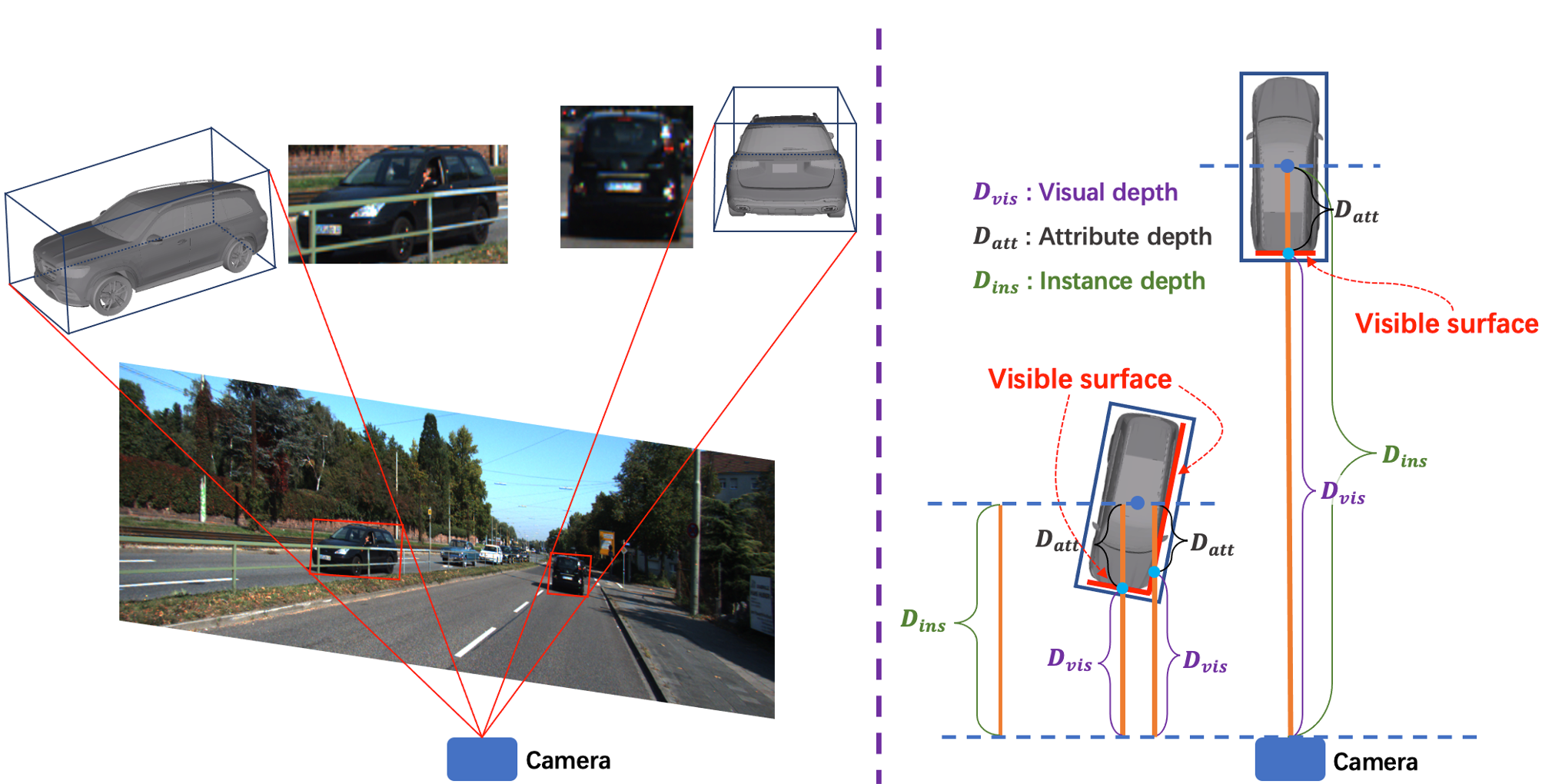}      
		\caption{
		We decouple the instance depth into visual depths and attribute depths due to the coupled nature of instance depth. 
		Please refer to the text for more details.
		Best viewed in color with zoom in.
				}
		\label{fig:intro}
\end{figure}

	Many previous works \cite{M3D,D4LCN,CaDDN} directly regress the instance depth.
	This manner does not consider the ambiguity brought by the instance depth itself.
	As shown in Figure \ref{fig:intro}, for the right object, its instance depth is the sum of car tail depth and half-length of the car, where the car length is ambiguous since both car's left and right sides are invisible.
	For the left object, except for the intuitive visible surface depth, the instance depth further depends on the car dimension and orientation.
	We can observe that the instance depth is non-intuitive.
	It requires the network to additionally learn instance inherent attributes on the instance depth head.
	Previous direct estimation and mediate optimization methods do not fully consider this coupled nature.
	Thus they lead to suboptimal performance on the instance depth estimation, showing less accurate results.

	Based on the analysis above, in this paper we propose to decouple the instance depth to \textbf{instance visual surface depth (visual depth)} and \textbf{instance attribute depth (attribute depth)}.
	We illustrate some examples in Figure \ref{fig:intro}.
	For each point (or small patch) on the object, the visual depth denotes the absolute depth towards the agent (car/robot) camera, and we define the attribute depth as the relative depth offset from the point (or small patch) to the object's 3D center.
	This decoupled manner encourages the network to learn different feature patterns of the instance depth.	 
	Visual depth on monocular imagery depends on objects' appearances and positions \cite{How_depth} on the image, which is affine-sensitive.
	By contrast, attribute depth highly relies on object inherent attributes ({\textit{e.g.}, dimensions and orientations}) of the object.
	It focuses on features inside the RoI, which is affine-invariant. (See Section \ref{sec:vis} and \ref{sec:att} for detailed discussion).
	Thus the attribute depth is independent of the visual depth, and decoupled instance depth allows us to use separate heads to extract different types of features for different types of depths.

	Specifically, for an object image patch, we divide it into $m\times n$ grids.
	Each grid can denote a small area on the object, with being assigned a visual depth and the corresponding attribute depth.
	Considering the occlusion and 3D location uncertainty, we use the uncertainty to denote the confidence of each depth prediction.
	At inference, every object can produce $m\times n$ instance depth predictions, thus we take advantage of them and associated uncertainties to adaptively obtain the final instance depth and confidence.

	Furthermore, prior works usually are limited by the diversity of data augmentation, due to the complexity of keeping alignment between 2D and 3D objects when enforcing affine transformation in a 2D image.
	Based on the decoupled instance depth, we show that our method can effectively perform data augmentation, including the way using affine transformation.
	It is achieved by the affine-sensitive and affine-invariant nature of the visual depth and attribute depth, respectively (See Section \ref{sec:aug}).
	To demonstrate the effectiveness of our method, we perform experiments on the widely used KITTI dataset.
	The results suggest that our method outperforms prior works with a significant margin.

	In summary, our contributions are listed as follows:

\begin{enumerate}
		\item We point out the coupled nature of instance depth.
			Due to the entangled features, the previous way of directly predicting instance depth is suboptimal.
			Therefore, we propose to decouple the instance depth into attribute depths and visual depths, which are independently predicted.
		\item We present two types of uncertainties to represent depth estimation confidence.
			 Based on this, we propose to adaptively aggregate different types of depths into the final instance depth and correspondingly obtain the 3D localization confidence.
		\item With the help of the proposed attribute depth and visual depth, we alleviate the limitation of using affine transformation in data augmentation for monocular 3D detection.
		\item Evaluated on KITTI benchmark, our method sets the new state of the art (SOTA).
			 Extensive ablation studies demonstrate the effectiveness of each component in our method. 
\end{enumerate}

\section{Related Work}

\subsection{LiDAR-based 3D Object Detection}
	LiDAR-based methods utilize precise LiDAR point clouds to achieve high performance.
	According to the representations usage, they can be categorized into point-based, voxel-based, hybrid, and range-view-based methods.
	Point-based methods \cite{qi2018frustum,shi2019pointrcnn,yang20203dssd} directly use raw point clouds to preserve fine-grained structures of objects. However, they usually suffer from high computational costs.
	Voxel-based methods \cite{zhou2018voxelnet,yan2018second,mao2021voxel,zheng2021se,Yin_2021_CVPR} voxelize the unordered point clouds into regular grids so that the CNNs can be easily applied. 
	These methods are more efficient, but voxelization introduces quantization errors, resulting in information loss.
	To explore advantages of different representations, some hybrid methods \cite{Chen2019fastpointrcnn,yang2019std,shi2020points,shi2020pv,noh2021hvpr} are proposed.
	They validate that combining point-based and voxel-based methods can achieve a better trade-off between accuracy and efficiency.
	Range-view-based methods \cite{bewley2020range,li2016vehicle,chai2021point,fan2021rangedet} organize point clouds in range view, which is a compact representation of point clouds.
	These methods are also computationally efficient but are under-explored.

\subsection{Monocular 3D Object Detection}
	Due to the low cost and setup simplicity, monocular 3D object detection is popular in recent years.  
	Previous monocular works can be roughly divided into image-only based and depth-map based methods.
	The pioneer method \cite{Mono3D} integrates different types of information such as segmentation and scene priors  for performing 3D detection.
	To learn spatial features, OFTNet \cite{OFTNet} projects 2D image features to the 3D space.
	M3D-RPN \cite{M3D} attempts to extract depth-aware features by using different convolution kernels on different image columns.
	Kinematic3D \cite{Kinematic3D} uses multi-frames to capture the temporal information by employing a tracking module.
	GrooMeD-NMS \cite{GrooMeD-NMS} develops a trainable NMS-step to boost the final performance.
	At the same period of time, many works fully take advantage of the scene and geometry priors \cite{MonoEF,MonoRCNN,PGD,Ground-Aware}. 
	Due to the ill-posed nature of monocular imagery, some works resort to using the dense depth estimation \cite{PseudoLiDAR,D4LCN,PatchNet,OCM3D,DDMP-3D,CaDDN,PCT}.
	With the help of estimated depth maps, RoI-10D \cite{RoI10D} use CAD models to augment training samples.
	Pseudo-LiDAR \cite{PseudoLiDAR,AM3D} based methods are also popular.
	They convert estimated depth maps to point clouds, then well-designed LiDAR 3D detectors can be directly employed.
	CaDDN \cite{CaDDN} predicts a categorical depth distribution, to precisely project depth-aware features to 3D space.
	In sum, benefiting from rapid developments of deep learning technologies,  monocular 3D detection has conducted remarkable progress.

\subsection{Estimation of Instance Depth}
	Most monocular works directly predict the instance depth.
	There are also some works that use auxiliary information to help the instance depth estimation in the post-processing.
	They usually take advantage of the geometry constraints and scene priors.
	The early work Deep3DBox \cite{Deep3DBBox} regresses object dimensions and orientations, the remaining 3D box parameters including the instance depth are estimated by 2D-3D box geometry constraints.
	This indirect way has poor performance because it does not fully use the supervisions.
	To use geometric projection constraints, RTM3D \cite{RTM3D} predicts nine perspective key-points (center and corners) of a 3D bounding box in the image space.
	Then the initially estimated instance depth can be optimized by minimizing projection errors.
	KM3D \cite{KM3D} follows this line and integrates this optimization into a end-to-end training process.
	More recently,  MonoFlex \cite{MonoFlex} also predicts the nine perspective key-points.
	In addition to the directly predicted instance depth, it uses the projection heights in pair key-points, using geometric relationships to produce new instance depths.
	MonoFlex develops an ensemble strategy to obtain the final instance depth.
	Differing from previous works, GUPNet \cite{GUPNet} proposes an uncertainty propagation method for the instance depth.
	It uses estimated object 3D dimensions and 2D height to obtain initial instance depth, with additionally predicting a depth bias to refine the instance depth.
	GUPNet mainly focuses on tackling the error amplification problem in the geometry projection process.
	MonoRCNN \cite{MonoRCNN} also introduces a distance decomposition based on the 2D height and 3D height.
	Such methods use geometric or auxiliary information to refine the estimated instance depth, while they do not fully use the coupled nature of the instance depth.

\section{Overview and Framework}
\subsubsection{Preliminaries}
	Monocular 3D detection takes an image captured by an RGB camera as input, predicting amodal 3D bounding boxes of objects in 3D space.
	These 3D boxes are parameterized by the 3D center location $(x,y,z)$, dimension $(h,w,l)$, and the orientation $(\theta)$.
	Please note, in the self-driving scenario, the orientation usually refers to the yaw angle, and the roll and pitch angles are zeros by default.
	Also, the ego-car/robot has been calibrated.

		\begin{figure}[t]
\centering 
				\includegraphics[width=0.9\linewidth]{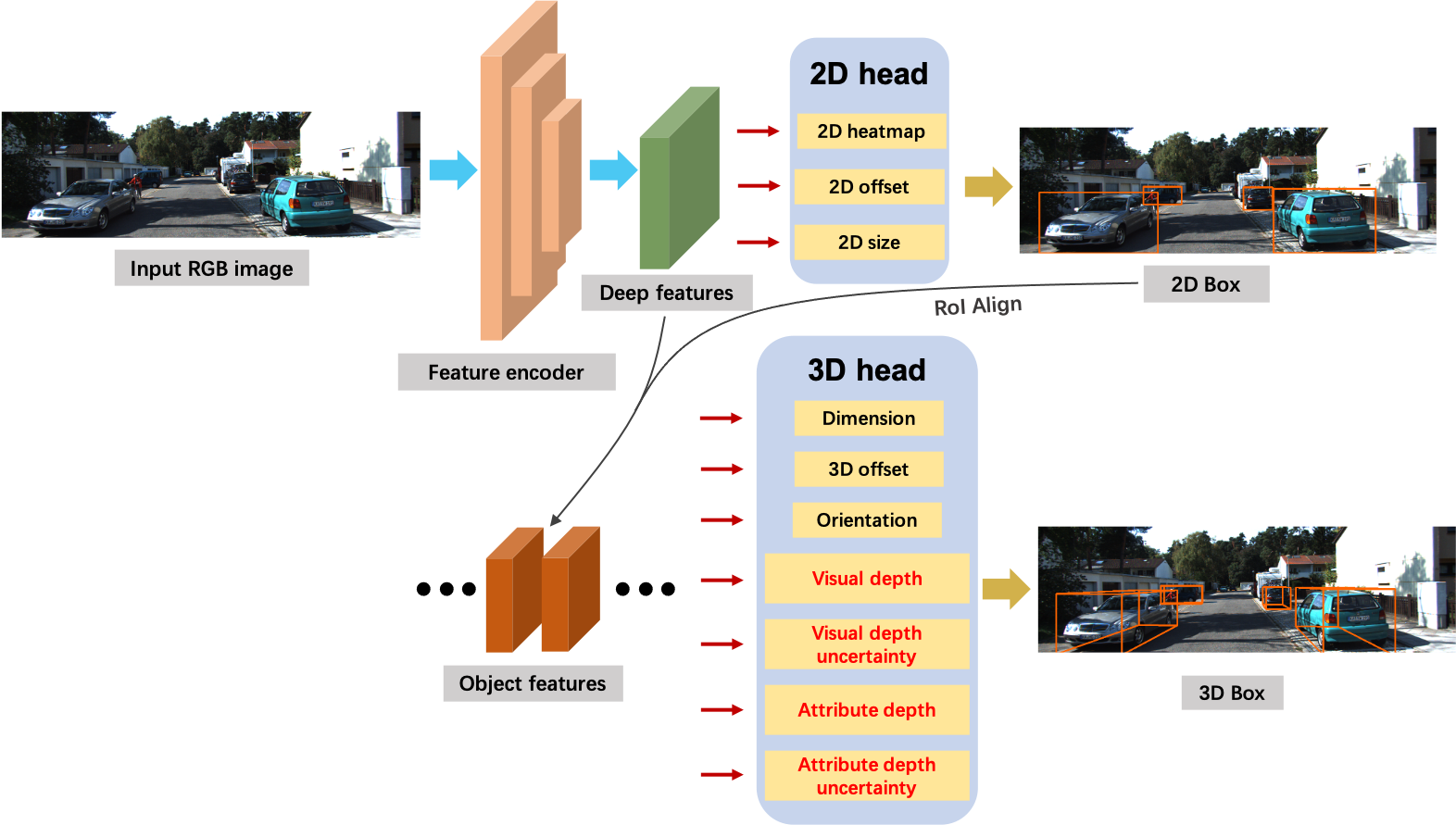}      
		\caption{
				Network framework. 
				The overall design follows GUPNet \cite{GUPNet}.
				The estimated 2D boxes are used to extract specific features for each object, followed by 3D box heads, which predict required 3D box parameters.
				The red parts in 3D heads denote the newly-proposed components.
				They are used to decouple the instance depth.
				}
		\label{fig:framework}
\end{figure}

	In this paragraph, we provide an overview and describe the framework.
	The overall framework is shown in Figure \ref{fig:framework}.
	First, the network takes an RGB image $\mathbf{I}\in\mathbb{R}^{H\times W\times3}$ as the input.
	After feature encoding, we have deep features $\mathbf{F}\in\mathbb{R}^{\frac{H}{4}\times \frac{W}{4}\times C}$, where $C$ is the channel number.
	Second, deep features $\mathbf{F}$ are fed into three 2D detection heads, namely, 2D heatmap $\mathbf{H}\in\mathbb{R}^{\frac{H}{4}\times \frac{W}{4}\times B}$, 2D offset $\mathbf{O}_{2d}\in\mathbb{R}^{\frac{H}{4}\times \frac{W}{4} \times 2}$, and 2D size $\mathbf{S}_{2d}\in\mathbb{R}^{\frac{H}{4}\times \frac{W}{4} \times 2}$, where $B$ is the number of categories.
	By combining such 2D head predictions, we can achieve 2D box predictions.
	Then, with 2D box estimates, single object features are obtained from deep features $\mathbf{F}$ by RoI Align.
	We have object features $\mathbf{F}_{obj}\in\mathbb{R}^{n\times7\times7\times C}$, where $7 \times 7$ is the RoI Align size and $n$ refers to the number of RoIs.
	Finally, these object features $\mathbf{F}_{obj}$ are fed into 3D detection heads to produce 3D parameters.
	Therefore, we have 3D box dimension  $\mathbf{S}_{3d}\in\mathbb{R}^{n \times 3}$, 3D center projection offset  $\mathbf{O}_{3d}\in\mathbb{R}^{n \times 2}$, orientation $\Theta\in\mathbb{R}^{n \times k \times 2}$ (we follow the multi-bin design \cite{Deep3DBBox} where $k$ is the bin number), visual depth $\mathbf{D}_{vis}\in\mathbb{R}^{n \times7\times7}$, visual depth uncertainty $\mathbf{U}_{vis}\in\mathbb{R}^{n \times7\times7}$, attribute depth $\mathbf{D}_{att}\in\mathbb{R}^{n \times7\times7}$, and attribute depth uncertainty $\mathbf{U}_{att}\in\mathbb{R}^{n \times7\times7}$.
	Using the parameters above, we can achieve the final 3D box predictions.
	In the following sections, we will detail the proposed method.

\section{Decoupled Instance Depth}  \label{sec:dec}
	We divide the RoI image patch into $7\times7$ grids, assigning each grid a visual depth value and an attribute depth value.
	We provide the ablation on the grid size for visual and attribute depth in experiments (See Section \ref{sec:grid_size} and Table \ref{tab:grid_size}).
	In the following, we first detail the two types of depths, followed by the decoupled-depth based data augmentation, then introduce the way of obtaining the final instance depth, and finally describe loss functions.

\subsection{Visual Depth}  \label{sec:vis}
	The visual depth denotes the physical depth of the object surface on the small RoI image grid. 
	For each grid, we define the visual depth as the average pixel-wise depth within the grid.
	If the grid is $1\times 1$ pixel, the visual depth is equal to the pixel-wise depth.
	Given that a pixel denotes the quantified surface of the object, we can regard visual depths as the general extension of pixel-wise depths.

	The visual depth in monocular imagery has an important property.
	For a monocular-based system, visual depth highly relies on the object's 2D box size (the faraway object appears small on the images and vice versa) and the position on the image (lower $v$ coordinates under image coordinate system indicate larger depths) \cite{How_depth}.
	Therefore, if we perform an affine transformation to the image, the visual depth should be correspondingly transformed, where the depth value should be scaled.
	We call this nature the affine-sensitive.

\subsection{Attribute Depth} \label{sec:att}
	The attribute depth refers to the depth offset from the visual surface to the object's 3D center.
	We call it attribute depth because it is more likely related to the object's inherent attributes.
	For example, when the car orientation is parallel to $z$-axis (depth direction) in 3D space, the attribute depth of the car tail is the car's half-length.
	By contrast, the attribute depth is car's half-width if the orientation is parallel to $x$-axis.
	We can see that the attribute depth depends on the object semantics and its inherent attributes. 
	In contrast to the affine-sensitive nature of visual depth, attribute depth is invariant to any affine transformation because object inherent characteristics will not change.
	We call this nature the affine-invariant.

	As described above, we use two separate heads to estimate the visual depth and attribute depth, respectively.
	The disentanglement of instance depth has several advantages: 
	(1) The object depth is decoupled in a reasonable and intuitive manner, thus we can more comprehensively and precisely represent the object; 
	(2) The network is allowed to extract different types of features for different types of depths, which facilitates the learning;
	(3) Benefitting from the decoupled depth, our method can effectively perform affine transformation based data augmentation, which is usually limited in previous works.

		\begin{figure}[t]
\centering 
				\includegraphics[width=0.9\linewidth]{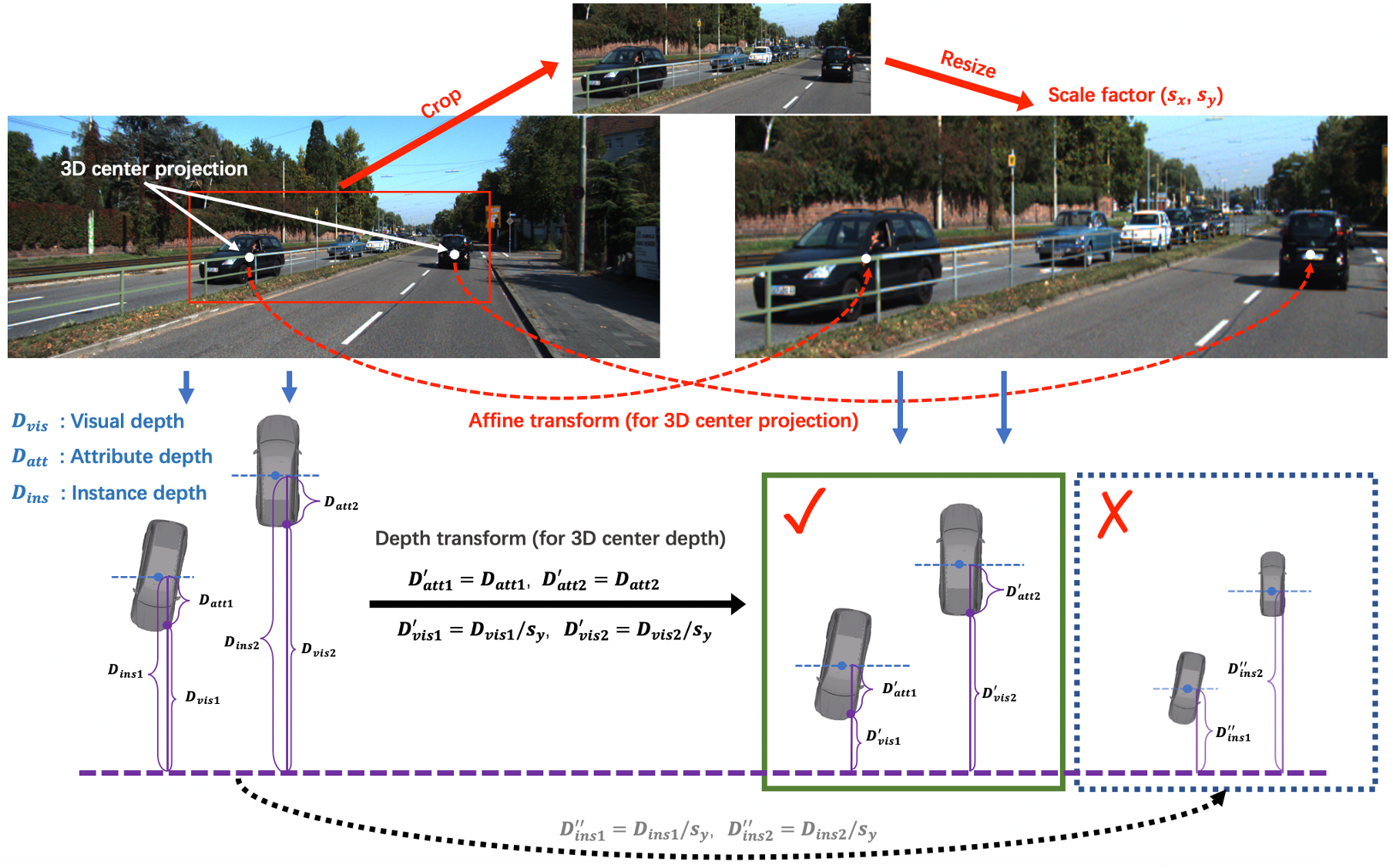}      
		\caption{
		Affine transformation based data augmentation.
		We do not change the object's inherent attributes, \textit{i.e.}, attribute depths, dimensions, and observation angles.
		The visual depth is scaled according to the 2D height scale factor.
		The 3D center projection is transformed together with the image affine transformation. 
				}
		\label{fig:data_aug}
\end{figure}

\subsection{Data Augmentation} \label{sec:aug}
	In monocular 3D detection, many previous works are limited by data augmentation.
	Most of them only employ photometric distortion and flipping transformation.
	Data augmentation using affine transformation is hard to be adopted because the transformed instance depth is agnostic. 
	Based on the decoupled depth, we point out that our method can alleviate this problem.

	We illustrate an example in Figure \ref{fig:data_aug}.
	Specifically, we add the random cropping and scaling strategy \cite{CenterNet} in the data augmentation. 
	The 3D center projection point on the image follows the same affine transformation process of the image.
	The visual depth is scaled by the scale factor along $y$-axis on the image because $d=\frac{f\cdot h_{3d}}{h_{2d}}$, where $f,h_{3d},h_{2d}$ is the focal length, object 3D height and 2D height, respectively.
	Conversely, the attribute depth keeps the same due to its affine-invariant nature.
	We do not directly scale the instance depth, as this manner will implicitly damage the attribute depth.
	Similarly, other inherent attributes of objects, \textit{i.e.}, the observation angle and the dimension, keep the same as original values.
	We empirically show that the data augmentation works well.
	We provide the ablations in experiments (See Section \ref{sec:abla_aug} and Table \ref{tab:data_aug}).

		\begin{figure}[t]
\centering 
				\includegraphics[width=0.8\linewidth]{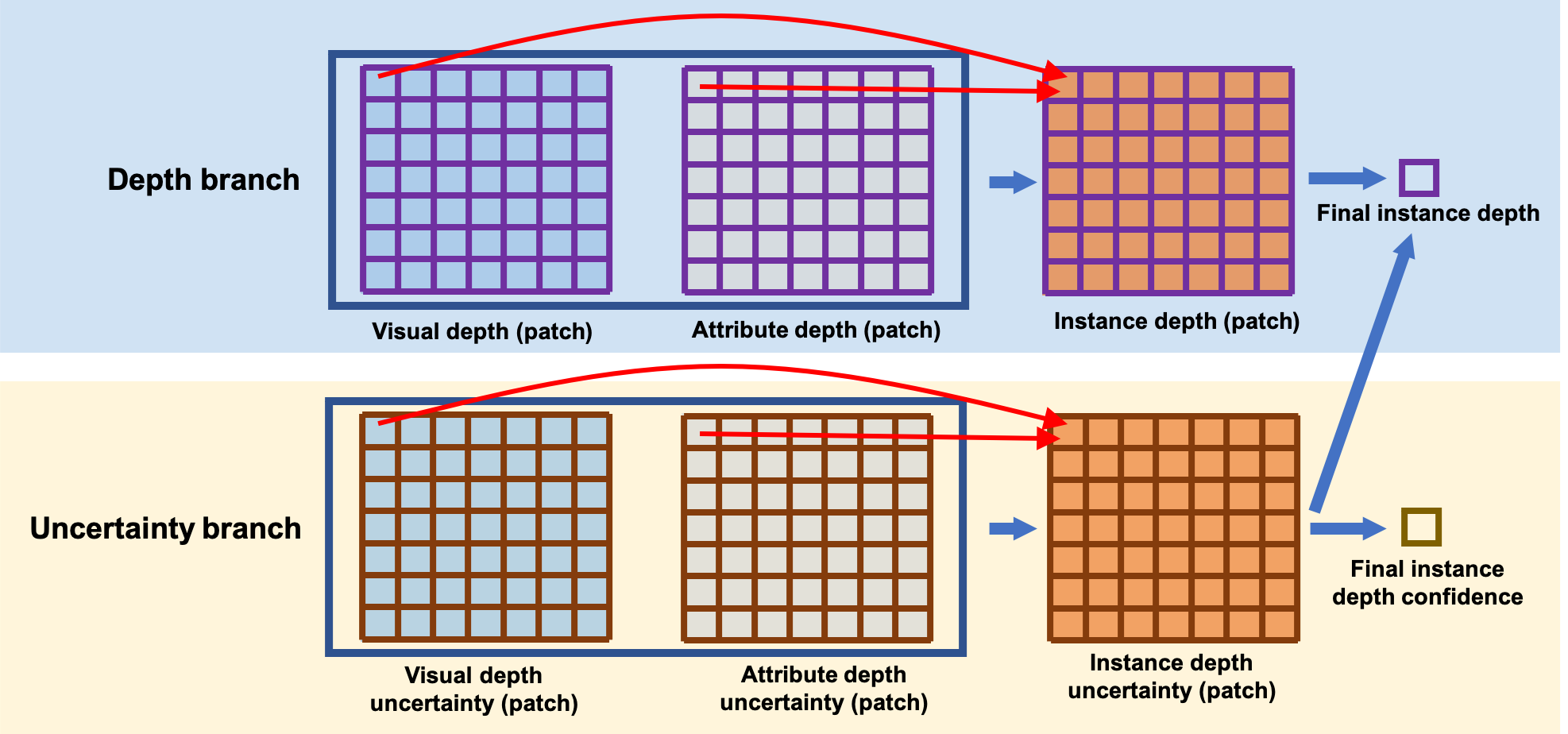}      
		\caption{
		Depth flow for an object. 
		We use the visual depth, attribute depth, and the associated uncertainty to obtain the final instance depth.
		}
		\label{fig:depth_flow}
\end{figure}

\subsection{Depth Uncertainty and Aggregation}
	The 2D classification score cannot fully express the confidence in monocular 3D detection because of the difficulty in 3D localization.
	Previous works \cite{GUPNet,AP40} use the instance depth confidence or 3D IoU loss, integrating with 2D detection confidence to represent the final 3D detection confidence.
	Given that we have decoupled the instance depth into visual depth and attribute depth, we can further decouple the instance depth uncertainty.
	Only when an object has low visual uncertainty and low attribute depth uncertainty simultaneously, the instance depth can have high confidence.
	
	Inspired by \cite{Uncern1,GUPNet}, we assume every depth prediction is a Laplace distribution.
	Specifically, for each visual depth $d_{vis}$ in $\mathbf{D}_{vis}\in\mathbb{R}^{n \times7\times7}$ and the corresponding uncertainty $u_{vis}$ in $\mathbf{U}_{vis}\in\mathbb{R}^{n \times7\times7}$,
	they follow the Laplace distribution $L(d_{vis},u_{vis})$.
	Similarly, the attribute depth distribution is $L(d_{att},u_{att})$, where $d_{att}$ in $\mathbf{D}_{att}\in\mathbb{R}^{n \times7\times7}$ and $u_{att}$ in $\mathbf{U}_{att}\in\mathbb{R}^{n \times7\times7}$.
	Therefore, the instance depth distribution derived by associated visual and attribute depth is $L(\tilde{d}_{ins},\tilde{u}_{ins})$, where $\tilde{d}_{ins}=d_{vis}+d_{att}$ and $\tilde{u}_{ins}=\sqrt{u_{vis}^2+u_{att}^2}$.
	Then we use $\mathbf{\tilde{D}}_{ins(patch)}\in\mathbb{R}^{n \times7\times7}$ and $\mathbf{\tilde{U}}_{ins(patch)}\in\mathbb{R}^{n \times7\times7}$ to denote the instance depth and the uncertainty on the RoI patch.
	We illustrate the depth flow process in Figure \ref{fig:depth_flow}.
	
	To obtain the final instance depth, we first convert the uncertainty to probability \cite{Uncern1,GUPNet}, which can be written as $\mathbf{P}_{ins(patch)}=exp(-\mathbf{\tilde{U}}_{ins(patch)})$, where $\mathbf{P}_{ins(patch)} \in \mathbb{R}^{n \times7\times7}$.
	Then we aggregate the instance depth on the patch to the final instance depth.
	For the $i^{th}$ object ($i=1,...,n$), we have:
\begin{equation}
	d_{ins} = \sum \frac{\mathbf{\tilde{D}}_{ins(patch)_i}  \mathbf{P}_{ins(patch)_i}}{\sum \mathbf{P}_{ins(patch)_i}}
	\label{equ:ins}
\end{equation}
	The corresponding final instance depth confidence is:
\begin{equation}
	p_{ins} =\sum (\frac{\mathbf{P}_{ins(patch)_i} }{\sum  \mathbf{P}_{ins(patch)_i}} \mathbf{P}_{ins(patch)_i} )
	\label{equ:prob_all}
\end{equation}
	Therefore, the final 3D detection confidence is $p = p_{2d} \cdot p_{ins}$, where $p_{2d}$ is the 2D detection confidence.

\subsection{Loss Functions}

\noindent
\textbf{2D detection part}:
	As shown in Figure \ref{fig:framework}, for the 2D object detection part, we follow the design in CenterNet \cite{CenterNet}.
	The 2D heatmap $\mathbf{H}$ aims to indicate the rough object center on the image.
	The size is $\frac{H}{4}\times\frac{W}{4}\times B$, where $H, W$ is the input image size and $B$ is the number of categories.
	The 2D offset $O_{2d}$ refers to the residual towards rough 2D centers, and the 2D size $S_{2d}$ denotes the 2D box height and width.
	Following CenterNet \cite{CenterNet}, we have loss functions $\mathcal{L}_{H},\mathcal{L}_{O_{2d}},\mathcal{L}_{S_{2d}}$, respectively.

~\\
\noindent
\textbf{3D detection part}:
	For the 3D object dimension, we follow the typical transformation and loss design \cite{M3D} $\mathcal{L}_{S_{3d}}$.
	For the orientation, the network predicts the observation angle and uses the multi-bin loss \cite{Deep3DBBox} $\mathcal{L}_{\Theta}$.
	Also, we use the 3D center projection on the image plane and the instance depth to recover the object's 3D location.
	For the 3D center projection, we achieve it by predicting the 3D projection offset to the 2D center.
	The loss function is: $\mathcal{L}_{O_{3d}}={\rm{Smooth}}L_1(O_{3d}, O_{3d}^\ast)$.
	We use $\ast$ to denote corresponding labels.
	As mentioned above, we decouple the instance depth into visual depth and attribute depth.
	The visual depth labels are obtained by projecting LiDAR points onto the image and the attribute depth labels are obtained by subtracting instance depth labels with visual depth labels.
	Combing with the uncertainty \cite{Uncern1,MonoPair}, the visual depth loss is: $\mathcal{L}_{D_{vis}}=\frac{\sqrt{2}}{u_{vis}}\|d_{vis}-d_{vis}^\ast\|+log(u_{vis})$, where $u_{vis}$ is the uncertainty.
	Similarly, we have attribute depth loss $\mathcal{L}_{D_{att}}$ and instance depth loss $\mathcal{L}_{D_{ins}}$.
	Among these loss terms, the losses concerning instance depth ($\mathcal{L}_{D_{vis}}$, $\mathcal{L}_{D_{att}}$, and $\mathcal{L}_{D_{ins}}$) play the most important role since they matter objects' localization in the 3D space. 
	We empirically set 1.0 for weights of all loss terms, and the overall loss is:
\begin{equation}
	\mathcal{L}=\mathcal{L}_{H}+\mathcal{L}_{O_{2d}}+\mathcal{L}_{S_{2d}}+\mathcal{L}_{S_{3d}}+\mathcal{L}_{\Theta}+\mathcal{L}_{O_{3d}}+\mathcal{L}_{D_{vis}}+\mathcal{L}_{D_{att}}+\mathcal{L}_{D_{ins}}
	\label{equ:loss_all}
\end{equation}

		  \begin{table}[h]
	  			\caption{
			Comparisons on KITTI testing set. 
			The \textcolor{red}{red} refers to the highest result and \textcolor{blue}{blue} is the second-highest result.
			Our method achieves state-of-the-art results.
			Note that DD3D\cite{DD3D} uses a large private dataset (DDAD15M), which includes 15M frames.
			}
			\label{tab:test}
								\resizebox*{1.0\linewidth}{0.475\textheight}{
	  \setlength{\tabcolsep}{1.2mm}
      \centering
						\begin{tabular}{l|c|c|ccc|ccc}
				\toprule   
				\multirow{2}{*}{Approaches} &\multirow{2}{*}{Venue} & \multirow{2}{*}{Runtime} & \multicolumn{3}{c|}{AP$_{BEV}$(IoU=0.7)$|\scriptstyle R_{40}$} & \multicolumn{3}{c}{AP$_{3D}$ (IoU=0.7)$|\scriptstyle R_{40}$} \\ 
				~&~& ~ & Easy & Moderate & Hard& Easy & Moderate & Hard\\ 
				\midrule         
				MonoGRNet \cite{MonoGRNet}& \textit{AAAI19}&400ms & 18.19  & 11.17  & 8.73 & 15.74  & 9.61  & 4.25 \\
				ROI-10D \cite{RoI10D}& \textit{CVPR19} &200ms & 9.78  & 4.91  & 3.74 & 4.32 & 2.02  & 1.46 \\ 
				MonoPSR \cite{MonoPSR}& \textit{CVPR19}  &200ms& 18.33 & 12.58 & 9.91 & 10.76 & 7.25 & 5.85 \\
				M3D-RPN \cite{M3D} & \textit{ICCV19}&160ms& 21.02  & 13.67  & 10.23  & 14.76  & 9.71  & 7.42 \\
				AM3D \cite{AM3D}& \textit{ICCV19}&400ms& 25.03 & 17.32 & 14.91 & 16.50 & 10.74 & 9.52 \\
				MonoPair \cite{MonoPair}& \textit{CVPR20}&60ms & 19.28 & 14.83 & 12.89 & 13.04 & 9.99 & 8.65\\
				D4LCN \cite{D4LCN}& \textit{CVPR20}&200ms & 22.51 & 16.02 & 12.55 & 16.65 & 11.72 & 9.51 \\
				RTM3D \cite{RTM3D} & \textit{ECCV20}&40ms& 19.17  & 14.20  &  11.99   & 14.41  & 10.34  &  8.77   \\ 
				PatchNet \cite{PatchNet}& \textit{ECCV20}&400ms& 22.97 & 16.86 & 14.97 & 15.68 & 11.12 & 10.17\\
				Kinematic3D \cite{Kinematic3D}& \textit{ECCV20}&120ms  & 26.69 & 17.52 & 13.10  & 19.07 & 12.72 & 9.17 \\
				Neighbor-Vote \cite{Neighbor-Vote}& \textit{MM21} &100ms & 27.39 & 18.65 & 16.54  & 15.57 & 9.90 & 8.89 \\
				Ground-Aware \cite{Ground-Aware}& \textit{RAL21}&50ms& 29.81& 17.98 & 13.08 & 21.65& 13.25 & 9.91  \\
				MonoRUn \cite{MonoRUn}& \textit{CVPR21}&70ms& 27.94 & 17.34 & 15.24  &19.65 & 12.30 & 10.58\\
				DDMP-3D \cite{DDMP-3D}& \textit{CVPR21} &180ms & 28.08 & 17.89 & 13.44  & 19.71 & 12.78 & 9.80\\
				Monodle \cite{Monodle} & \textit{CVPR21}&40ms & 24.79 & 18.89 & 16.00  & 17.23 & 12.26 & 10.29\\
				CaDDN \cite{CaDDN} & \textit{CVPR21}&630ms & 27.94 & 18.91 & 17.19  & 19.17 & 13.41 & 11.46 \\
				GrooMeD-NMS \cite{GrooMeD-NMS}& \textit{CVPR21} &120ms& 26.19 & 18.27 & 14.05 & 18.10 & 12.32 & 9.65\\
				MonoEF \cite{MonoEF}& \textit{CVPR21} &30ms& 29.03 & 19.70 & 17.26   & 21.29 & 13.87 & 11.71\\
				MonoFlex \cite{MonoFlex}& \textit{CVPR21}  &35ms & 28.23 & 19.75 & 16.89 & 19.94 & 13.89 & 12.07  \\
				MonoRCNN \cite{MonoRCNN}& \textit{ICCV21}&70ms &  25.48 & 18.11 & 14.10  & 18.36 & 12.65 & 10.03\\
				AutoShape \cite{AutoShape}& \textit{ICCV21}&40ms &  30.66 & 20.08 & 15.95  & 22.47 & 14.17 & 11.36\\
				GUPNet \cite{GUPNet}  & \textit{ICCV21} &34ms & 30.29 & 21.19 & 18.20 & 22.26 & 15.02 & 13.12  \\
				PCT \cite{PCT} & \textit{NeurIPS21} &45ms & 29.65 & 19.03 & 15.92 & 21.00 & 13.37  & 11.31\\
				MonoCon \cite{MonoCon} & \textit{AAAI22} &26ms &\bf \textcolor{blue}{31.12} & \bf \textcolor{blue}{22.10} &\bf \textcolor{blue}{19.00} & \bf \textcolor{blue}{22.50} &\bf \textcolor{red}{16.46} &\bf \textcolor{red}{13.95}  \\
				\textit{DD3D \cite{DD3D}} & \textit{ICCV21} &- &{\textit{32.35}} &{\textit{23.41}} & {\textit{20.42}} &{\textit{23.19}} &{\textit{16.87}} & {\textit{14.36}}  \\
				\midrule 
				{\bf DID-M3D (ours)}&\multirow{2}{*}{\textit{ECCV22}}&\multirow{2}{*}{40ms} &\bf \textcolor{red}{32.95} &\bf \textcolor{red}{22.76}  &\bf \textcolor{red}{19.83}  &\bf \textcolor{red}{24.40}  & \bf \textcolor{blue}{16.29}  & \bf \textcolor{blue}{13.75} \\
				 Improvements &~ & ~ & +1.83 & +0.66 & +0.83 & +1.90 & -0.17 & -0.20 \\ 
				\bottomrule 
			\end{tabular}}
  \end{table}

\section{Experiments}

\subsection{Implementation Details}
	We conduct experiments on 2 NVIDIA RTX 3080Ti GPUs with batch size 16.
	We use the PyTorch framework \cite{pytorch}.
	We train the network with 200 epochs and employ the Hierarchical Task Learning (HTL) \cite{GUPNet} training strategy.
	The Adam optimizer is used with the initial learning rate $1e-5$.
	The learning rate increases to $1e-3$ in the first 5 epochs by employing the linear warm-up strategy and decays in epoch 90 and 120 with rate 0.1. 
	We set $k$ 12 in the multi-bin orientation $\Theta$.
	For the backbone and head, we follow the design in \cite{GUPNet,DLA34}.
	Inspired by CaDDN \cite{CaDDN}, we project LiDAR point clouds onto the image frame to create sparse depth maps and then depth completion \cite{hu2021penet} is performed to generate depth values at each pixel in the image.
	Considering the space limitation, we provide more experimental results and discussion in the supplementary material.

\subsection{Dataset and Metrics}
	Following the commonly adopted setup in previous works \cite{M3D,D4LCN,GrooMeD-NMS,AutoShape}, we perform experiments on the widely used KITTI \cite{KITTI2012} 3D detection dataset.
	The KITTI dataset provides 7,481 training samples and 7,518 testing samples, where training sample labels are publicly available and the labels of testing samples keep secret in the KITTI website, which are only used for online evaluation and ranking.
	To conduct ablations, previous work further divides the 7,481 samples into a new training set with 3,712 samples and a \textit{val} set with 3,769 samples.
	This data split \cite{3DOP} is widely adopted by most previous works.
	Additionally, KITTI divides objects into the \textit{easy}, \textit{moderate}, and \textit{hard} level according to the object 2D box height (related to the depth), occlusion and truncation levels.
	For evaluation metrics, we use the suggested $AP_{40}$ metric \cite{AP40} under the two core tasks, \textit{i.e.}, 3D and bird's-eye-view (BEV) detection.

 		\begin{figure}[t]
\centering 
				\includegraphics[width=0.9\linewidth]{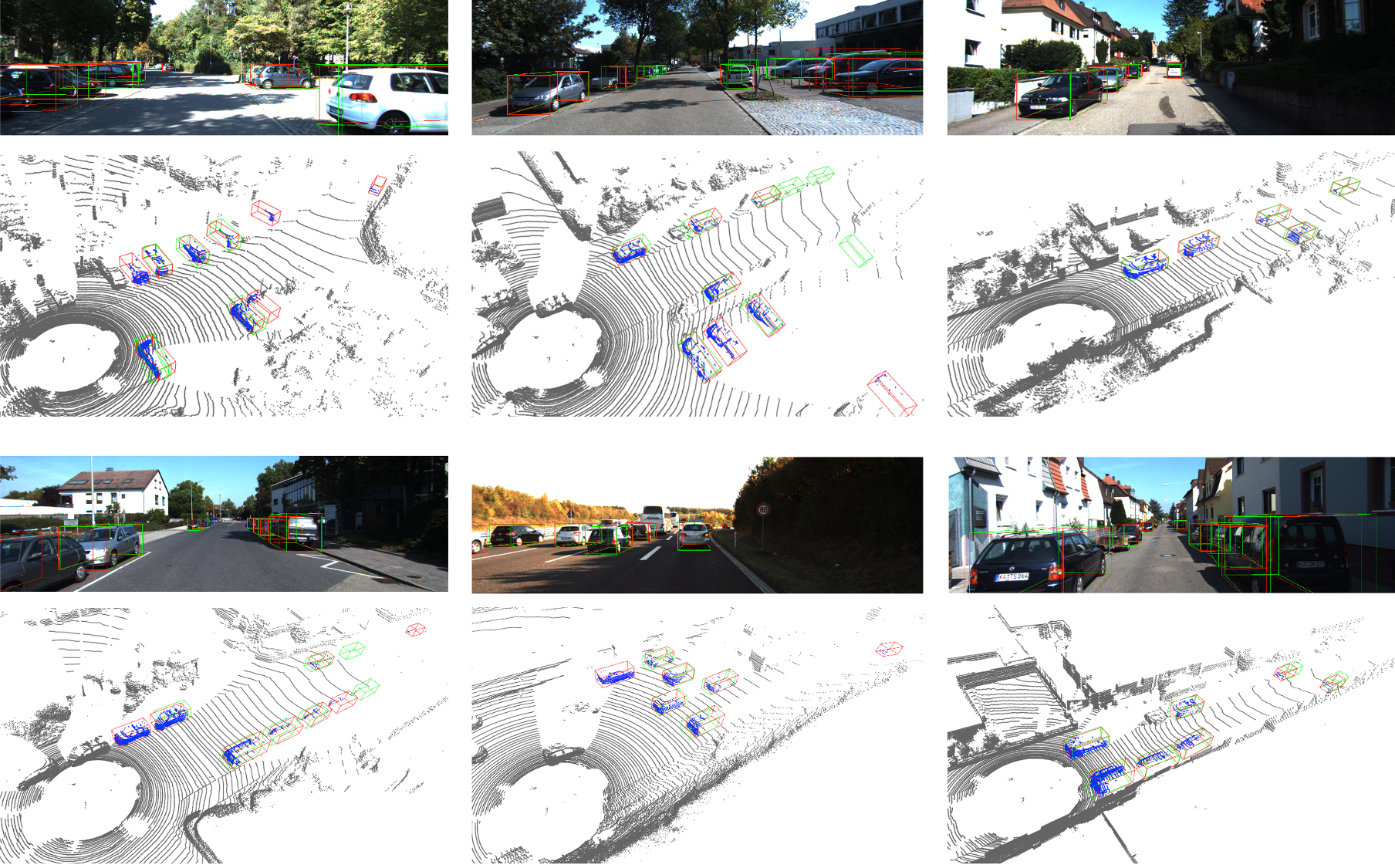}      
		\caption{
		Qualitative results on KITTI \textit{val} set. \textcolor{red}{Red:} ground-truth 3D boxes; \textcolor{green}{Green:} our predictions. 
		We can observe that the model conducts accurate 3D box predictions.
		Best viewed in color with zoom in.
		}
		\label{fig:qua}
\end{figure}

\subsection{Performance on KITTI Benchmark}
	We compare our method (DID-M3D) with other methods in KITTI test set, which is the official benchmark for monocular 3D object detection.
	The results are shown in Table \ref{tab:test}.
	We can see that our method achieves a new state-of-the-art.
	For example, compared to GUPNet \cite{GUPNet} (\textit{ICCV21}), we boost the performance from 21.19/15.02 to 22.26/16.29 under the moderate setting.
	As for PCT \cite{PCT} (\textit{NeurIPS21}), we exceed it with 3.23/2.92 AP under the moderate setting, which is a significant improvement.
	When compared to the recent method MonoCon\cite{MonoCon} (\textit{AAAI22}), our method still shows better performance on all BEV metrics and a 3D metric.
	Also, the runtime of our method is comparable to other real-time methods. 
	Such results validate the superiority of our method.
	Additionally, to demonstrate the generalizability on other categories, we perform experiments on cyclist and pedestrian categories.
	As shown in Table \ref{tab:ped}, our method brings obvious improvements to the baseline (without employing the proposed components).
	The results suggest that our method works well for other categories.
	
	Moreover, we provide qualitative results on the RGB image and 3D space in Figure \ref{fig:qua}.
	We can observe that for most simple cases (\textit{e.g.}, close objects without occlusion and truncation), the model predictions are quite precise.
	However, for the heavily occluded, truncated, or faraway objects, the orientation or instance depth is less accurate.
	This is a common dilemma for most monocular works due to the limited information in monocular imagery.
	In the supplementary material, we will provide more experimental results and make detailed discussions on failure cases.

\subsection{Ablation Study}
	To investigate the impact of each component in our method, we conduct detailed ablation studies on KITTI.
	Following the common practice in previous works, all ablation studies are evaluated in the \textit{val} set on the car category.
	
\subsubsection{Decoupled Instance Depth.}
	We report the results in Table \ref{tab:main_abala}.
	Experiment (a) is the baseline using the direct instance depth prediction.
	To make fair comparisons, for the baseline, we also employ the grid design (Experiment (b)).
	Similar to our method, it means that the network also produce $7 \times 7$ instance depth predictions for every object, which are all supervised in training and averaged at inference.
	Then, we decouple the instance depth into visual depths and attribute depths (Experiment (b) $\rightarrow$ (c)), this simple modification improves the accuracy significantly.
	This result indicates the network performs suboptimally due to the coupled nature of instance depth, demonstrating our viewpoint.
	From Experiment (c) $\rightarrow$ (d,e), we can see that the depth uncertainty brings improvements, because the uncertainty stabilizes the training of depth, benefiting the network learning.
	When simultaneously enforcing both two types of uncertainty, the performance is further boosted.
	Please note, the decoupled instance depth is the precondition of decoupled uncertainty.
	Given that the two types of depth uncertainty are achieved, we can obtain the final instance depth uncertainty (Experiment (f) $\rightarrow$ (g)).
	This can be regarded as the 3D location confidence.
	It is used to combine with the original 2D detection confidence, which brings obvious improvements.
	Finally, we can use the decoupled depth and corresponding uncertainties to adaptively obtain the final instance depth (Experiment (h)), while previous experiments use the average value on the patch.
	We can see that this design enhances the performance.
	In summary, by using the decoupled depth strategy, we improve the baseline  performance from 16.79/11.24 to 22.76/16.12 (Experiment (b) $\rightarrow$ (h)).
	It is an impressive result.
	Overall, the ablations validate the effectiveness of our method.

	\begin{table}[h]
\centering
        		\caption{ 
		Ablation for decoupled instance depth. 
		{``Dec."}: decoupled;
		{``ID."}: instance depth;
		{``$u_{vis}$"}: visual depth uncertainty;
		{``$u_{att}$}": attribute depth uncertainty;
		{``Conf."}: confidence;
		{``AA."}: adaptive aggregation.
		}
		\label{tab:main_abala}
									\resizebox*{1.0\linewidth}{0.19\textheight}{
\begin{tabular}{c|c|c|cc|c|c|ccc}
				\toprule  
				\multirow{2}{*}{Experiments} &\multirow{2}{*}{Grid} & \multirow{2}{*}{Dec. ID.} & \multirow{2}{*}{$u_{vis}$} &\multirow{2}{*}{$u_{att}$}& \multirow{2}{*}{ID. Conf.}& \multirow{2}{*}{ID. AA.} &  \multicolumn{3}{c}{AP$_{BEV}$/AP$_{3D}$ (IoU=0.7)$|\scriptstyle R_{40}$} \\ 
				~&~&~&~ &~&~ &~  &  Easy & Moderate & Hard \\ 
				\midrule    
				   (a)&~&~&~ &~&~ &~ &19.84/14.07 & 16.25/11.20 & 14.13/9.97 \\
				   (b)& \checkmark&~&~ &~&~ &~ &21.01/14.67 & 16.79/11.24 & 15.41/10.18 \\
				   \midrule
				   (c)& \checkmark& \checkmark &~& ~ &~&~ & 22.98/16.95 & 18.72/13.24 & 16.57/11.23 \\
				    (d)& \checkmark&\checkmark &\checkmark& ~ &~&~ &25.13/16.85 & 19.84/13.45 & 17.19/12.03\\
				    (e)& \checkmark&\checkmark &~& \checkmark &~&~ &24.83/17.29 & 19.66/13.50 & 17.06/11.48 \\
				    (f)& \checkmark&\checkmark &\checkmark& \checkmark &~&~ &25.23/18.14 & 20.06/13.91 & 17.63/12.52 \\
				    (g)& \checkmark&\checkmark &\checkmark& \checkmark &\checkmark&~ &29.34/21.51 & 21.53/15.57 & 18.55/12.84\\
				    (h)& \checkmark&\checkmark &\checkmark& \checkmark &\checkmark &\checkmark & \bf 31.10/22.98 &\bf 22.76/16.12 &\bf 19.50/14.03  \\
					\bottomrule 
			\end{tabular}}
\end{table}

\subsubsection{Affine Transformation Based Data Augmentation.} \label{sec:abla_aug}
	In this paragraph we aim to understand the effect of affine transformation based data augmentation.
	The comparisons are shown in Table \ref{tab:data_aug}.
	We can see that the method obviously benefits from affine-based data augmentation. 
	Note that the proper depth transformation is very important.
	When enforcing affine-based data augmentation, the visual depth should be scaled while the attribute depth should not be changed due to their affine-sensitive and affine-invariant nature, respectively.
	If we change the attribute depth without scaling visual depth, the detector even performs worse than the one without affine-based data augmentation ($AP_{3D}$ downgrades from 12.76 to 12.65).
	It is because this manner misleads the network in the training with incorrect depth targets.
	After revising visual depth, the network can benefit from the augmented training samples, boosting the performance from 19.05/12.76 to  21.74/15.48 AP under the moderate setting.
	We can see that the improper visual depth can result in larger impacts compared to the improper attribute depth on the final performance, as the visual depth has a larger value range.
	Finally, we obtain the best performance when employing the proper visual depth and attribute depth transformation strategy.

    	\begin{table}[h]
\centering
        		\caption{ 
		Ablation for affine transformation based data augmentation. 
		{``Aff. Aug."} in the table denotes the affine-based data augmentation.
		}
		\label{tab:data_aug}
							\resizebox*{0.8\linewidth}{0.14\textheight}{
\begin{tabular}{c|p{1.7cm}<{\centering}p{1.7cm}<{\centering}|ccc}
				\toprule  
				\multirow{2}{*}{w/ Aff. Aug.} & \multirow{2}{*}{Scaled $d_{vis}$} &\multirow{2}{*}{Scaled $d_{att}$} &  \multicolumn{3}{c}{AP$_{BEV}$/AP$_{3D}$ (IoU=0.7)$|\scriptstyle R_{40}$} \\ 
				~ &~ & ~ &   Easy & Moderate & Hard \\ 
				\midrule    
				   ~ &~ & ~&  25.63/17.61 & 19.05/12.76 & 16.34/11.21 \\
				     \checkmark &~ & ~ & 26.97/18.98&  19.80/14.33 & 17.71/12.00  \\
				   \checkmark &~ & \checkmark&  22.23/15.27 & 18.98/12.65 & 16.52/10.75 \\
				 \checkmark &\checkmark & \checkmark &28.67/21.43 &21.74/15.48 & 18.76/13.47\\
				   \midrule
				\checkmark & \checkmark & ~ & \bf 31.10/22.98 &\bf 22.76/16.12 &\bf 19.50/14.03\\ 
					\bottomrule 
			\end{tabular}}
\end{table}

\subsubsection{Grid Size for Visual and Attribute Depth.} \label{sec:grid_size}
	As described in Section \ref{sec:dec}, we divide the RoI image patch into $m \times m$ grids, where each grid has a visual depth and an attribute depth.
	This paragraph investigates the impact brought by the grid size.
	When increasing grid size $m$, visual depths and attribute depths are becoming fine-grained.
	This tendency makes visual depth more intuitive, which is close to the pixel-wise depth.
	However, the fine-grained grid will lead to suboptimal performance in terms of learning object attributes since the attributes focus on the overall object.
	It indicates that there exists a trade-off.
	Therefore, we perform ablations on the grid size $m$, as shown in Table \ref{tab:grid_size}.
	We achieve the best performance when $m$ is set to 7.

  	\begin{table}[t]
\centering
 		\caption{ 
		Ablation for the grid size on visual depth and attribute depth. 
		}
		\label{tab:grid_size}
					\resizebox*{0.5\linewidth}{0.17\textheight}{
\begin{tabular}{c|ccc}
				\toprule  
				\multirow{2}{*}{Grid size} &  \multicolumn{3}{c}{AP$_{BEV}$/AP$_{3D}$ (IoU=0.7)$|\scriptstyle R_{40}$} \\ 
				~ &  Easy & Moderate & Hard \\ 
				\midrule    
				   $1 \times 1$&26.11/19.39 &18.76/13.01 &16.00/11.34\\
				    $3 \times 3$&27.03/19.73 &19.91/14.33&18.10/12.25\\
				   $5 \times 5$&29.20/21.36 & 21.53/15.13 &18.61/12.53\\
				   $9 \times 9$& 30.21/21.78 &22.28/14.98& 18.93/12.47\\
				   $13 \times 13$& 29.88/21.67 &21.97/15.29& 18.96/12.72\\
				  $19 \times 19$&28.20/20.36 &21.53/15.13&18.61/12.53\\
				   \midrule
				 \bm{$7 \times 7$}&\bf 31.10/22.98 &\bf 22.76/16.12 &\bf 19.50/14.03 \\ 
					\bottomrule 
			\end{tabular}}
\end{table}

    		  \begin{table}[h]
	  			\caption{
			Comparisons on pedestrian and cyclist categories on KITTI \textit{val} set under IoU criterion 0.5.
			Our method brings obvious improvements to the baseline.
			}
			\label{tab:ped}
			\resizebox*{1.0\linewidth}{0.09\textheight}{
	  \setlength{\tabcolsep}{1.5mm}
      \centering
						\begin{tabular}{l|ccc|ccc}
				\toprule   
				\multirow{2}{*}{Approaches} & \multicolumn{3}{c|}{Pedestrian, AP$_{BEV}$/AP$_{3D}$$|\scriptstyle R_{40}$} &\multicolumn{3}{c}{Cyclist, AP$_{BEV}$/AP$_{3D}$$|\scriptstyle R_{40}$} \\ 
				~ & Easy & Moderate & Hard& Easy & Moderate & Hard\\ 
				\midrule
				Baseline &5.97/5.13 &4.75/3.88  & 3.87/3.05 & 5.11/4.27 & 2.68/2.46 & 2.50/2.17   \\
				Baseline+Ours &\bf 8.86/7.27 & \bf 7.01/5.87 & \bf 5.46/4.89  & \bf 6.13/5.54 &\bf 3.09/2.59 &\bf 2.67/2.49 \\
				\bottomrule 
			\end{tabular}}
  \end{table}

\section{Conclusion}
	In this paper, we point out that the instance depth is coupled by visual depth clues and object inherent attributes.
	Its entangled nature makes it hard to be precisely estimated with the previous direct method.
	Therefore, we propose to decouple the instance depth into visual depths and attribute depths.
	This manner allows the network to learn different types of features for instance depth.
	At the inference stage, the instance depth is obtained by aggregating visual depth, attribute depth, and associated uncertainties.
	Using the decoupled depth, we can effectively perform affine transformation based data augmentation on the image, which is usually limited in previous works. 
	Finally, extensive experiments demonstrate the effectiveness of our method.

\section*{Acknowledgments}
This work was supported in part by The National Key Research and Development Program of China (Grant Nos: 2018AAA0101400), in part by The National Nature Science Foundation of China (Grant Nos: 62036009, U1909203, 61936006, 61973271), in part by Innovation Capability Support Program of Shaanxi (Program No. 2021TD-05).

%
%
\bibliographystyle{splncs04}
\bibliography{egbib}

\clearpage
\appendix
\begin{center}
	\textbf{\Large Appendix}
\end{center}

 \renewcommand\thesection{\Alph{section}}
 
   \definecolor{c0}{HTML}{663300}
 \definecolor{c1}{HTML}{c91f37}
 \definecolor{c1_2}{HTML}{336699}
 \definecolor{c2}{HTML}{FF9900}
 \definecolor{c3}{HTML}{4b5cc4}
 \definecolor{c4}{HTML}{0eb83a}
  \definecolor{c5}{HTML}{ef7a82}
  \definecolor{c6}{HTML}{9933FF}
  
    \definecolor{c7}{HTML}{33CC33}
    \definecolor{c8}{HTML}{FF33CC}

Considering the space limitation of the main text, we provided more results and discussion in this supplementary material, which is organized as follows:
\begin{itemize}
	\item Section {\ref{sec:waymo}}: results on Waymo dataset.
	\item Section {\ref{sec:main_abla}}: detailed ablation analysis and discussion.
		\begin{itemize}
  \item[$\bullet$] Section {\ref{sec:dec}}: decoupled instance depth.
  \item[$\bullet$] Section {\ref{sec:grid}}: grid design.
  \item[$\bullet$] Section {\ref{sec:aff}}: affine-based data augmentation.
  \item[$\bullet$] Section {\ref{sec:unc}}: instance depth uncertainty.
  \item[$\bullet$] Section {\ref{sec:agg}}: instance depth aggregation.
		\end{itemize}
	\item Section {\ref{sec:qua_res}}: qualitative results.
		\begin{itemize}
	\item Section {\ref{sec:qua_att}}: analysis on attribute depth and visual depth uncertainty.
  	\item Section {\ref{sec:qua_3d}}: more qualitative results of 3D box predictions.
	\item Section {\ref{sec:qua_fail}}: failure cases and discussion.
		\end{itemize}
\end{itemize}

\section{Results on Waymo Dataset}\label{sec:waymo}
	We perform experiments on Waymo dataset \cite{waymo}, which is a large-scale modern dataset for self-driving.
	It contains 798 sequences for training and 202 sequences for validation.
	We use the same data split strategy proposed in CaDDN \cite{CaDDN}.
	The processed training dataset includes approximately 50K training samples.
	We show the results in Table \ref{tab:waymo}.
	We can see that our method performs the best on most metrics.
	It further validates the effectiveness of the proposed method.

		 		\begin{table}[h]
				
        \centering
        \scriptsize
        \setlength{\tabcolsep}{1.5mm}
        			        \caption{
			        Results on Waymo \textit{val} set. 
			        \textit{``Runtime$^*$"} in the table refers to the runtime reported on KITTI.
			        DID-M3D performs the best.
			        }
			\label{tab:waymo}    
			\begin{tabular}{l|c|c|cccc}
				\toprule  
				  \multirow{2}{*}{\tabincell{l}{Methods}}&  \multirow{2}{*}{\tabincell{c}{Venue}}&\multirow{2}{*}{\tabincell{c}{Runtime$^*$}}&  \multicolumn{4}{c}{3D mAP/mAPH} \\ 
				~ &~& ~&Overall & 0$-$30m & 30$-$50m & 50m$-$$\infty$ \\   
				\midrule
				\multicolumn{7}{c}{Under Level 1 (IoU=0.5)} \\
				\midrule
			 PatchNet  \cite{PatchNet}& \textit{ECCV20}  & 488ms & 2.92/2.74 & 10.03/9.75  & 1.09/0.96  & 0.23/0.18 \\ 
			 CaDDN  \cite{CaDDN} & \textit{CVPR21} & 630ms &  \bf \textcolor{blue}{17.54/17.31} & \bf \textcolor{red}{45.00/44.46}  & \bf \textcolor{blue}{9.24/9.11} & 0.64/0.62\\ 
			 PCT \cite{PCT}  & \textit{NeurIPS21}  & 445ms&  4.20/4.15 & 14.70/14.54  & 1.78/1.75  & 0.39/0.39\\ 
			 MonoJSG  \cite{MonoJSG} & \textit{CVPR22} & 42ms &  5.65/5.47 & 20.86/20.26  & 3.91/3.79  & \bf \textcolor{blue}{0.97/0.92} \\ 
			 {\bf DID-M3D} & \textit{ECCV22} & 40ms &\bf \textcolor{red}{20.66/20.47} & \bf \textcolor{blue}{40.92/40.60} &\bf \textcolor{red}{15.63/15.48} &\bf \textcolor{red}{5.35/5.24} \\ 		
				\midrule
				\multicolumn{7}{c}{Under Level 2 (IoU=0.5)} \\
				\midrule
			 PatchNet \cite{PatchNet} & \textit{ECCV20} & 488ms &  2.42/2.28 &10.01/9.73 & 1.07/0.94 & 0.22/0.16 \\ 
			 CaDDN \cite{CaDDN} & \textit{CVPR21} & 630ms&  \bf \textcolor{blue}{16.51/16.28} & \bf \textcolor{red}{44.87/44.33}  & \bf \textcolor{blue}{8.99/8.86}  & 0.58/0.55\\ 
			 PCT \cite{PCT}  & \textit{NeurIPS21} & 445ms &  4.03/3.99 & 14.67/14.51 & 1.74/1.71 & 0.36/0.35\\ 
			 MonoJSG \cite{MonoJSG} & \textit{CVPR22} & 42ms &  5.34/5.17 & 20.79/20.19 & 3.79/3.67 & \bf \textcolor{blue}{0.85/0.82} \\ 
			 {\bf DID-M3D} & \textit{ECCV22} & 40ms  &\bf \textcolor{red}{19.37/19.19} & \bf \textcolor{blue}{40.77/40.46} &\bf \textcolor{red}{15.18/15.04} &\bf \textcolor{red}{4.69/4.59} \\ 					
				\bottomrule 
			\end{tabular}
      \end{table}

\section{Detailed Ablation Analysis and Discussion}\label{sec:main_abla}
	To better understand the effect of each component in our method, we perform more detailed ablation studies.
	The results are shown in Table \ref{tab:main_abala}.
	We perform 5 groups of experiments, \textit{i.e.}, experiment ([a,b,c], [d,e,f,g], [h,i,j,k], [l,m,n,o], [p,q,r,s,t,u]), to compare the decoupled depth with the original prediction.
	We also extend our grid design, affine transformation based data augmentation, depth uncertainty, and depth aggregation to the baseline for comprehensive comparisons.  
	For comparison convenience, we copy the results from experiment (e, g, i, k) to experiment (m, o, q, s), respectively.

\subsection{Decoupled Instance Depth}\label{sec:dec}
	As shown in Table \ref{tab:main_abala}, for every group of experiments, we investigate the effect of the decoupled instance depth.
	We can easily see that the decoupled design consistently improves the overall performance with a significant margin.
	For example, on the naive baseline, the decoupled design boosts the performance from 13.43/8.70 to 16.49/10.94 (experiment b$\rightarrow$c) under the moderate setting.
	When employing other strategies, it also improves the AP from 19.82/14.47 to 22.76/16.12 (experiment t$\rightarrow$u) under the moderate setting.
	These improvements validate its effectiveness.

\begin{table}[h]
\centering
        \scriptsize
        		\caption{ 
		Detailed ablation studies. 
		\textcolor{c0}{``E.": experiments};
		\textcolor{c1}{``Dec. ID.": decoupled instance depth};
		\textcolor{c1_2}{``G.": grid};
		\textcolor{c2}{``Aff. Aug.": affine-based data augmentation};
		\textcolor{c3}{``Tr. ID.": transformed instance depth};
		\textcolor{c4}{``ID. U.": instance depth uncertainty};
		\textcolor{c5}{``ID. C.": instance depth confidence};
		\textcolor{c6}{``ID. AA.": instance depth adaptive aggregation}.
		The transformed instance depth (\textcolor{c3}{``Tr. ID."}) refers to the depth transformation in the affine-based data augmentation.
		}
		\label{tab:main_abala}
\begin{tabular}{c|c|c|c|c|c|c|c|ccc}
				\toprule  
				\multirow{2}{*}{\textcolor{c0}{E.}} & \multirow{2}{*}{\textcolor{c1}{Dec. ID.}} &\multirow{2}{*}{\textcolor{c1_2}{G.}} & \multirow{2}{*}{\textcolor{c2}{Aff. Aug.}} & \multirow{2}{*}{\textcolor{c3}{Tr. ID.}} & \multirow{2}{*}{\textcolor{c4}{ID. U.}} & \multirow{2}{*}{\textcolor{c5}{ID. C.}}& \multirow{2}{*}{\textcolor{c6}{ID. AA.}} &  \multicolumn{3}{c}{AP$_{BEV}$/AP$_{3D}$ (IoU=0.7)$|\scriptstyle R_{40}$} \\ 
				~&~&~&~ &~&~ &~ & ~ &  Easy & Moderate & Hard \\ 
				\midrule    
				   (a)&~&~&~ &~&~ &~ & ~&15.52/10.55  & 12.43/8.54 & 11.42/7.14 \\
				   (b)& ~&\checkmark&~ &~&~ &~ & ~&16.23/10.94 & 13.43/8.70 & 11.88/7.98 \\
				    (c)& \checkmark&\checkmark&~ &~&~ &~ & ~&\bf 19.86/13.13 &\bf 16.49/10.94 &\bf 14.40/9.89   \\
				   \midrule
				   (d)& ~&~&\checkmark &~&~ &~ & ~& 16.91/11.21  & 13.20/9.08 & 12.07/8.16 \\
				   (e)& \checkmark&\checkmark&\checkmark &~&~ &~ & ~&17.98/12.11 &15.34/10.72 & 14.11/9.11 \\
				   (f)& ~&~&\checkmark &\checkmark&~ &~ & ~&19.84/14.07 & 16.25/11.20 & 14.13/9.97 \\
				   (g)& \checkmark&\checkmark&\checkmark &\checkmark&~ &~ & ~&\bf 22.98/16.95 &\bf 18.72/13.24 &\bf 16.57/11.23  \\
				   \midrule
				    (h)& ~&~&\checkmark &\checkmark&\checkmark &~ & ~& 21.75/15.94 & 17.87/12.64 & 15.45/11.16 \\
				    (i)& \checkmark&\checkmark&\checkmark &\checkmark&\checkmark &~ & ~&25.23/18.14 & 20.06/13.91 & 17.63/12.52 \\
				    (j)& ~&~&\checkmark &\checkmark&\checkmark &\checkmark & ~ & 25.56/18.91 & 19.68/14.13 & 16.94/12.32 \\
				    (k)& \checkmark&\checkmark&\checkmark &\checkmark&\checkmark &\checkmark & ~&\bf 29.34/21.51 &\bf 21.53/15.57 &\bf 18.55/12.84\\
				    \midrule
				    (l)& ~&\checkmark&\checkmark &~&~ &~ & ~&17.60/12.14 & 14.09/9.37 & 12.36/8.42 \\
				 (m)& \checkmark&\checkmark&\checkmark &~&~ &~ & ~&17.98/12.11 &15.34/10.72 & 14.11/9.11  \\
				   (n)& ~&\checkmark&\checkmark &\checkmark&~ &~ & ~&21.01/14.67 & 16.79/11.24 & 15.41/10.18 \\
				   (o)& \checkmark&\checkmark&\checkmark &\checkmark&~ &~ & ~&\bf 22.98/16.95 &\bf 18.72/13.24 &\bf 16.57/11.23 \\
				   \midrule
				    (p)& ~&\checkmark&\checkmark &\checkmark&\checkmark &~ & ~&22.76/16.56 & 18.40/12.99 & 16.15/11.06 \\
				    (q)& \checkmark&\checkmark&\checkmark &\checkmark&\checkmark &~ & ~&25.23/18.14 & 20.06/13.91 & 17.63/12.52 \\
				    (r)& ~&\checkmark&\checkmark &\checkmark&\checkmark &\checkmark & ~&26.61/19.44 & 19.72/14.44 & 16.85/12.13 \\
				    (s)& \checkmark&\checkmark&\checkmark &\checkmark&\checkmark &\checkmark & ~&29.34/21.51 & 21.53/15.57 & 18.55/12.84\\
				   (t)& ~&\checkmark&\checkmark &\checkmark&\checkmark &\checkmark & \checkmark& 26.72/19.48 & 19.82/14.47 & 16.93/12.15 \\
				   (u)& \checkmark&\checkmark&\checkmark &\checkmark&\checkmark &\checkmark & \checkmark& \bf 31.10/22.98 & \bf 22.76/16.12 &\bf 19.50/14.03 \\
					\bottomrule 
			\end{tabular}
\end{table}	

\subsection{Grid Design}\label{sec:grid}
	Most previous monocular works produce a single instance depth prediction.
	Our method divides the RoI into grids to decouple the instance depth.
	The grid design produces multiple predictions, which may be unfair to the single prediction.
	Therefore, we extend the grid design to the baseline for fair comparisons.
	We can see that the baseline benefits from this grid design with slight improvements.
	The naive baseline obtains 1.0/0.24 AP improvements (experiment a$\rightarrow$b)  under the moderate setting.
	However, when the network is equipped with other useful components, gains from the grid design are weakened (\textit{e.g.}, 19.68/14.13$\rightarrow$19.82/14.47 (experiment j$\rightarrow$t) under the moderate setting).

\subsection{Affine-based Data Augmentation}\label{sec:aff}
	We extend the affine-based data augmentation to the baseline detector,
	Please note, in this process we directly scale the instance depth, as the baseline uses the direct instance depth prediction.
	When using affine-based data augmentation, the baseline benefits from it a lot (\textit{e.g.}, 12.43/8.54$\rightarrow$16.25/11.20 (experiment a$\rightarrow$f) under the moderate setting).
	Even if without a correct instance depth transformation, the performance is still boosted (12.43/8.54$\rightarrow$13.20/9.08 (experiment a$\rightarrow$d) under the moderate setting), which can be attributed to the improvements on the robustness for a simple baseline. 
	
	On the other hand, for our decoupled manner, the instance depth is decoupled in a more intuitive and reasonable way, thus incorrect depth transformation damages the accuracy (downgrading from 16.49/10.94 to 15.34/10.72 (experiment c$\rightarrow$e) under the moderate setting).
	With the correct depth transformation, our method obtains significant improvements via the affine-based data augmentation (2.23/2.30 AP gains (experiment c$\rightarrow$g) under the moderate setting).

\subsection{Instance depth Uncertainty}\label{sec:unc}
	We also investigate the impact brought by the uncertainty.
	The uncertainty can stabilize the training process as it allows the network to learn more reasonable objects.
	We can observe that this strategy brings improvements for both the coupled and the decoupled manner (\textit{e.g.}, 16.25/11.20$\rightarrow$17.87/12.64 (experiment f$\rightarrow$h) and 18.72/13.24$\rightarrow$20.06/13.91 (experiment g$\rightarrow$i) under the moderate setting).
	Intuitively, once the network has learned the depth uncertainty, we can use it to represent the confidence of instance depth estimation.
	This depth confidence can also express the 3D location confidence, which is used to combine with 2D detection confidence as the final 3D detection confidence.
	Based on this, the performance is further boosted  (\textit{e.g.}, 1.81/1.49 gains (experiment h$\rightarrow$j) and 1.47/1.66 gains (experiment i$\rightarrow$k) under the moderate setting).

\subsection{Instance Depth Aggregation}\label{sec:agg}
	In previous experiments, for the grid design, the final instance depth is the average value in the RoI grids.
	Given that every instance depth estimation in each grid has the corresponding uncertainty, we can use uncertainties in the grids to adaptively obtain the final instance depth.
	Thus we perform experiments r$\rightarrow$t (for the original coupled manner) and s$\rightarrow$u (for our decoupled manner).
	Interestingly, we can observe that this depth aggregation in the grid cannot bring significant improvements to the original coupled manner (only 0.1/0.03 gains under the moderate setting).
	This is because all direct instance depth estimates are very close.
	By contrast, for our decoupled manner, we obtain obvious improvements (1.23/0.55 gains under the moderate setting).
	It indicates that different parts of the objects can produce different features, to conduct different instance depth predictions with associated uncertainties.
	This experiment further validates the presence of the coupled nature in instance depth and demonstrates the effectiveness of our method.

 \begin{figure}[h]
\centering 
				\includegraphics[width=1.0\linewidth]{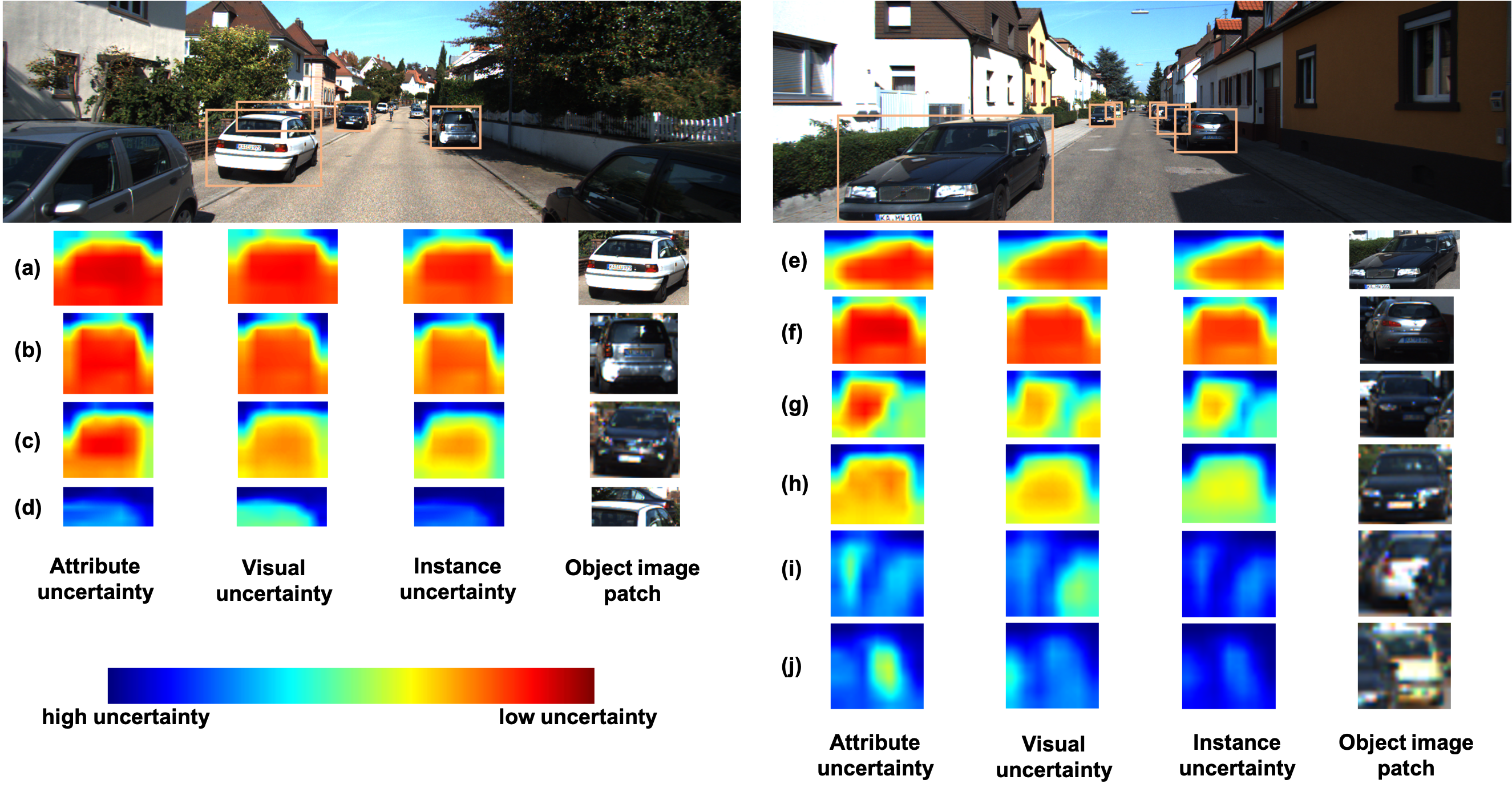}      
		\caption{
		Attribute depth and visual depth uncertainty.
		For simple objects, they show similar distributions.
		While they have different characteristics for difficult objects.
		Please refer to Section \ref{sec:qua_att} for more discussion.
		Best viewed in color with zoom in.
		}
		\label{fig:qua_att}
\end{figure}

\section{Qualitative Results}\label{sec:qua_res}

\subsection{Analysis on Attribute Depth and Visual Depth Uncertainty}\label{sec:qua_att}
	To better understand how the attribute depth and visual depth work, we illustrate their uncertainties on two typical scenes, as shown in Figure \ref{fig:qua_att}.
	$\bullet$ First, we can easily see that all background areas of RoIs have high uncertainties.
	It is expected because the background area does not have important clues to the estimation of foreground objects, and its visual depths and attribute depths are hard to predict.
	$\bullet$ Second, regarding simple objects such as objects (a, b, e, f), their attribute depth uncertainties and visual depth uncertainties have similar distributions, since both two types of depths are easy to estimate.
	$\bullet$ Third, for the far objects (c, h), we know that the visual depth is less confident than the attribute depth.
	This is reasonable as the object texture is obvious, and the network can be confident in estimating attribute depths.
	By contrast, predicting absolute visual depths for far objects is difficult.
	$\bullet$ Finally, concerning the occluded objects (d, g, i, j), we can observe that visual depths and attribute depths have different interest areas.
	Visual depths mainly focus on closer objects because of the prediction simplicity. 
	In contrast to visual depths, attribute depths are more interested in the target object, which even is heavily occluded (objects (i, j)).
	When the object is nearly invisible, attribute depths will give all areas high uncertainties (object (d)).
	

\subsection{More Qualitative Results of 3D Box Predictions}\label{sec:qua_3d}
	We provide more qualitative results in Figure \ref{fig:qua}.
	For better visualization, the 3D box predictions are drawn in the 3D space and on the RGB image simultaneously.
	We can see that our method works well in most scenes, which proves its effectiveness and robustness.

 \begin{figure}[h]
\centering 
				\includegraphics[width=1.0\linewidth]{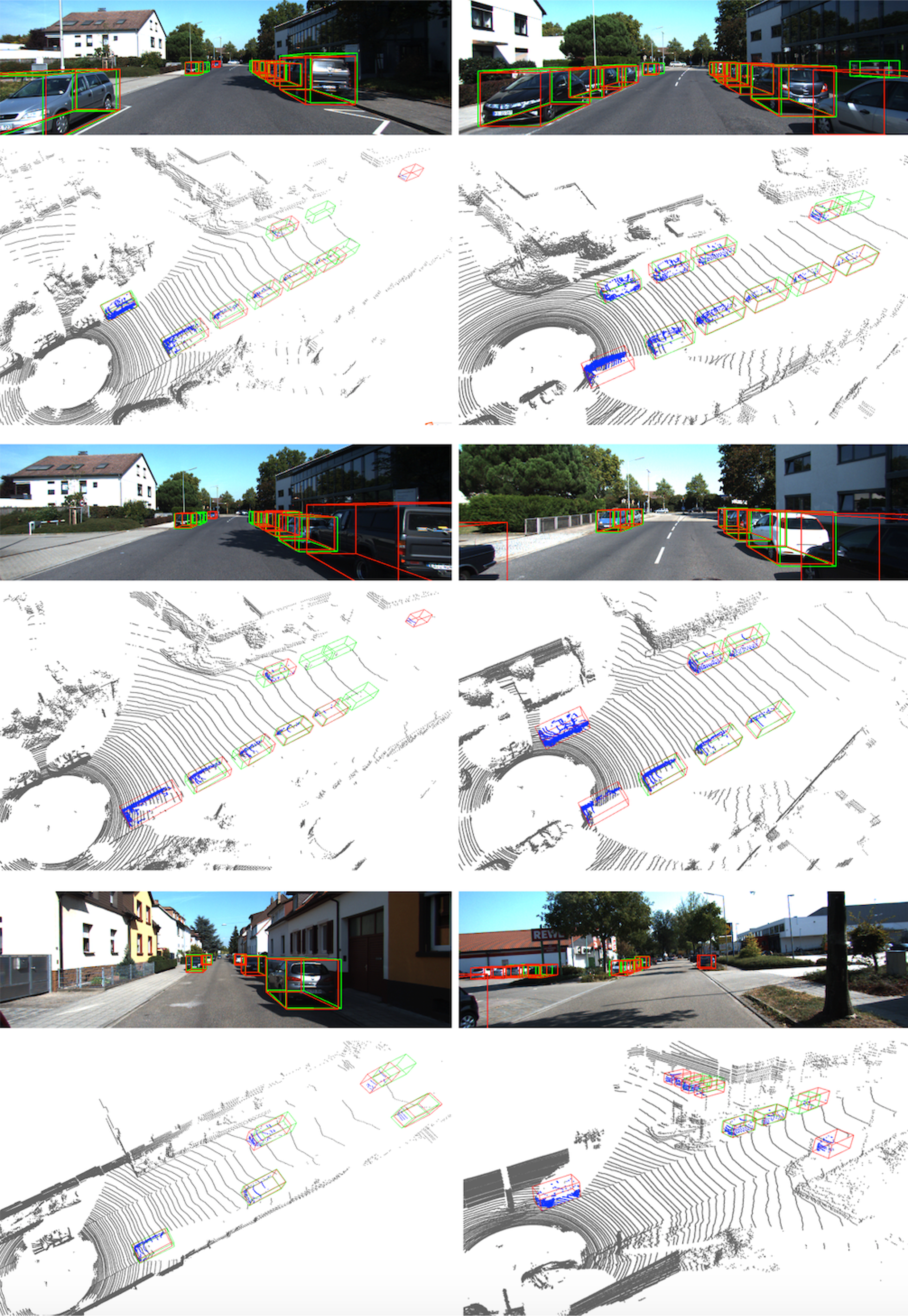}      
		\caption{
		Qualitative results on KITTI \textit{val} set. \textcolor{red}{Red:} ground-truth 3D boxes; \textcolor{c7}{Green:} our predictions. 
		We can observe that most 3D box predictions are quite accurate.
		The LiDAR point clouds are only used for visualization.
		Best viewed in color with zoom in.
		}
		\label{fig:qua}
\end{figure}
\clearpage

\subsection{Failure Cases and Discussion}\label{sec:qua_fail}
\begin{figure}[h]
\centering 
				\includegraphics[width=1.0\linewidth]{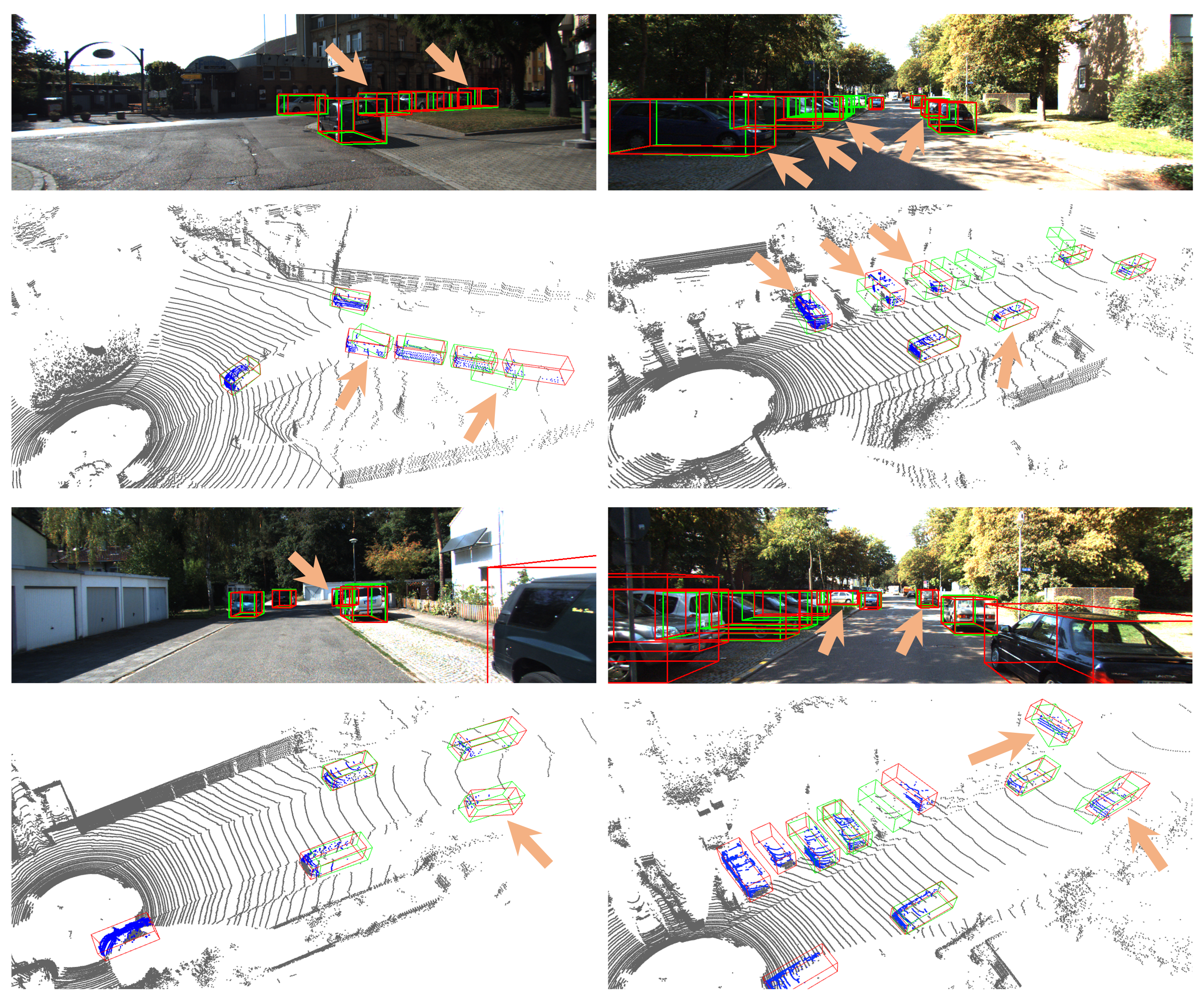}      
		\caption{
		Failure cases on KITTI \textit{val} set. \textcolor{red}{Red:} ground-truth 3D boxes; \textcolor{c7}{Green:} our predictions. 
		We use arrows to indicate failure cases.
		We can see that failure objects usually are faraway, occluded, truncated, or suffer from poor illumination.
		The LiDAR point clouds are only used for visualization.
		Best viewed in color with zoom in.
		}
		\label{fig:qua_fail}
\end{figure}

	This paragraph aims to investigate failure cases in our method.
	We show some examples in Figure \ref{fig:qua_fail}.
	These failure cases can be roughly divided into four categories, \textit{i.e.}, faraway, occluded, truncated, and poor illumination.
	Faraway objects are very difficult to be precisely predicted due to the dramatically decreased information along with the depth in monocular imagery.
	Some occluded and truncated objects provide few valuable clues on the image to infer their locations and orientations.
	As for objects in poor illumination, their texture features are weakened, thus bringing difficulty in estimating their 3D boxes. 
	In future works, we will further explore these failure cases, to mitigate their adverse impacts.

\end{document}